\documentclass[sigconf]{acmart}

\usepackage{algorithmic}
\usepackage{algorithm}

\def\BibTeX{{\rm B\kern-.05em{\sc i\kern-.025em b}\kern-.08emT\kern-.1667em\lower.7ex\hbox{E}\kern-.125emX}}
    
\copyrightyear{2019}
\acmYear{2019}
\acmConference[WWW 2019]{The Web Conference 2019}{May 13-17, 2019}{San Francisco,
  CA, USA}
\begin{document}

%
\title{Joint Entity Linking with Deep Reinforcement Learning}

%
\author{Zheng Fang}
\affiliation{%
  \institution{Institute of Information Engineering, Chinese Academy of Sciences \& School of Cyber Security, University of Chinese Academy of Sciences}
}
\email{fangzheng@iie.ac.cn}

\author{Yanan Cao}
\authornote{Corresponding author.}
\affiliation{%
  \institution{Institute of Information Engineering, Chinese Academy of Sciences}
}
\email{caoyanan@iie.ac.cn}

\author{Dongjie Zhang}
\affiliation{%
  \institution{Institute of Information Engineering, Chinese Academy of Sciences}
}
\email{zhangdongjie@iie.ac.cn}

\author{Qian Li}
\affiliation{%
  \institution{University of Technology Sydeney}
}
\email{Qian.Li@uts.edu.au}

\author{Zhenyu Zhang}
\affiliation{%
  \institution{Institute of Information Engineering, Chinese Academy of Sciences}
}
\email{zhangzhenyu1996@iie.ac.cn}

\author{Yanbing Liu}
\affiliation{%
  \institution{Institute of Information Engineering, Chinese Academy of Sciences}
}
\email{liuyanbing@iie.ac.cn}

%

%
\begin{abstract}
Entity linking is the task of aligning mentions to corresponding entities in a given knowledge base. Previous studies have highlighted the necessity for entity linking systems to capture the global coherence. However, there are two common weaknesses in previous global models. First, most of them calculate the pairwise scores between all candidate entities and select the most relevant group of entities as the final result. In this process, the consistency among wrong entities as well as that among right ones are involved, which may introduce noise data and increase the model complexity. Second, the cues of previously disambiguated entities, which could contribute to the disambiguation of the subsequent mentions, are usually ignored by previous models. To address these problems, we convert the global linking into a sequence decision problem and propose a reinforcement learning model which makes decisions from a global perspective. Our model makes full use of the previous referred entities and explores the long-term influence of current selection on subsequent decisions. We conduct experiments on different types of datasets, the results show that our model outperforms state-of-the-art systems and has better generalization performance.
\end{abstract}

%
%


%

%
\maketitle

\section{Introduction}
Entity Linking (EL), which is also called Entity Disambiguation (ED), is the task of mapping mentions in text to corresponding entities in a given knowledge Base (KB). This task is an important and challenging stage in text understanding because mentions are usually ambiguous, i.e., different named entities may share the same surface form and the same entity may have multiple aliases. EL is key for information retrieval (IE) and has many applications, such as knowledge base population (KBP), question answering (QA), etc.

Existing EL methods can be divided into two categories: local model and global model. Local models concern mainly on contextual words surrounding the mentions, where mentions are disambiguated independently. These methods are not work well when the context information is not rich enough. Global models take into account the topical coherence among the referred entities within the same document, where mentions are disambiguated jointly. Most of previous global models \cite{GaneaH17, TitovL18a, NguyenFRHGS16} calculate the pairwise scores between all candidate entities and select the most relevant group of entities. However, the consistency among wrong entities as well as that among right ones are involved, which not only increases the model complexity but also introduces some noises. For example, in Figure 1, there are three mentions "France", "Croatia" and "2018 World Cup", and each mention has three candidate entities. Here, "France" may refer to \emph{French Republic}, \emph{France national basketball team} or \emph{France national football team} in KB. It is difficult to disambiguate using local models, due to the scarce common information in the contextual words of "France" and the descriptions of its candidate entities. Besides, the topical coherence among the wrong entities related to \emph{basketball team} (linked by an orange dashed line) may make the global models mistakenly refer "France" to \emph{France national basketball team}. So, how to solve these problems?

\begin{figure*}[t]
\centering
\includegraphics[width=5.6in]{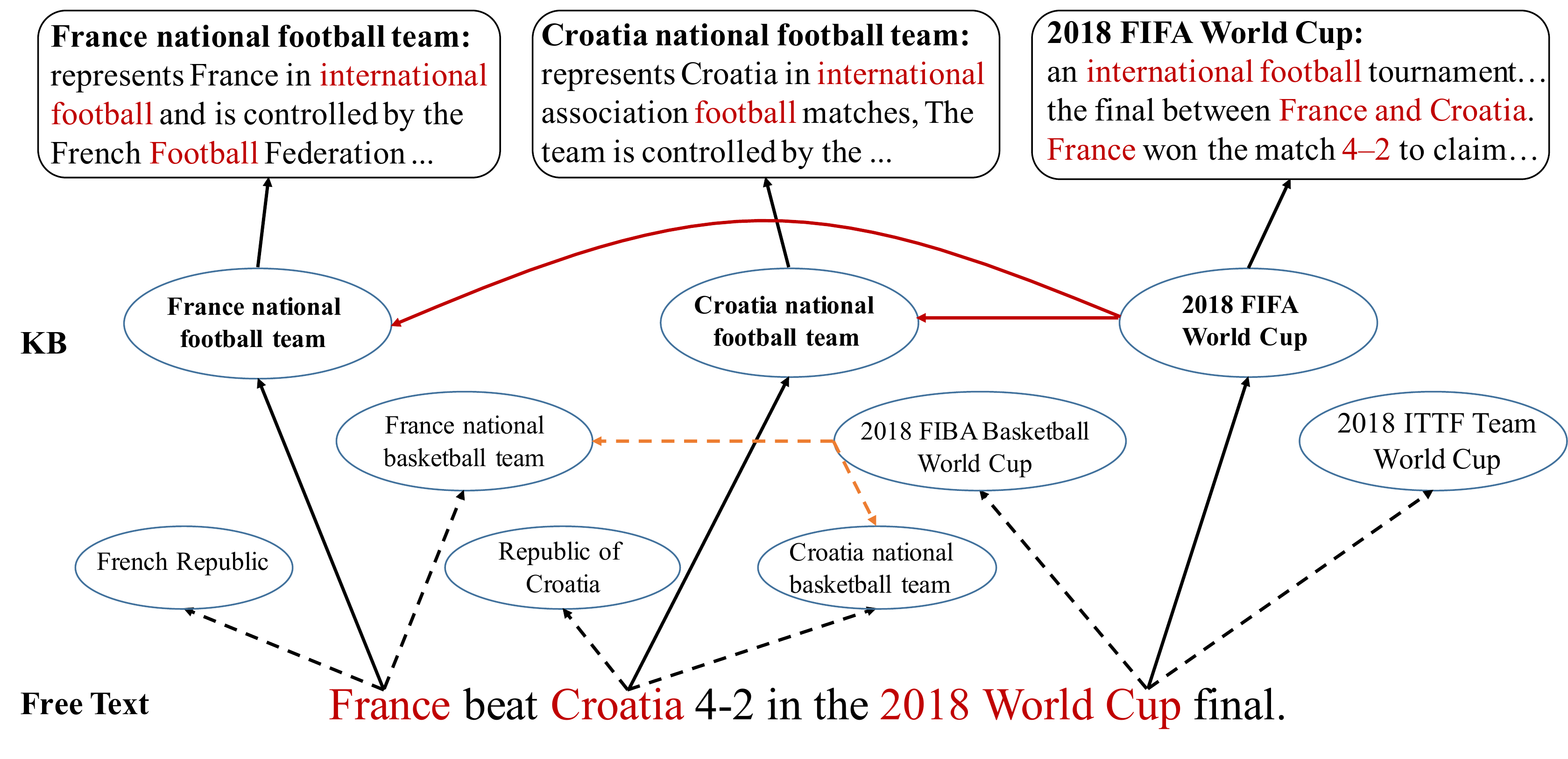}
\caption{Illustration of mentions in the free text and their candidate entities in the knowledge base. Solid black lines point to the correct target entities corresponding to the mentions and to the descriptions of these correct target entities. Solid red lines indicate the consistency between correct target entities and the orange dashed lines denote the consistency between wrong candidate entities.}
\end{figure*}

We note that, mentions in text usually have different disambiguation difficulty according to the quality of contextual information and the topical coherence. Intuitively, if we start with mentions that are easier to disambiguate and gain correct results, it will be effective to utilize information provided by previously referred entities to disambiguate subsequent mentions. In the above example, it is much easier to map "2018 World Cup" to \emph{2018 FIFA World Cup} based on their common contextual words "France", "Croatia", "4-2". Then, it is obvious that "France" and "Croatia" should be referred to the national football team because football-related terms are mentioned many times in the description of \emph{2018 FIFA World Cup}.

Inspired by this intuition, we design the solution with three principles: (i) utilizing local features to rank the mentions in text and deal with them in a sequence manner; (ii) utilizing the information of previously referred entities for the subsequent entity disambiguation; (iii) making decisions from a global perspective to avoid the error propagation if the previous decision is wrong.

In order to achieve these aims, we consider global EL as a sequence decision problem and proposed a deep reinforcement learning (RL) based model, RLEL for short, which consists of three modules: Local Encoder, Global Encoder and Entity Selector. For each mention and its candidate entities, Local Encoder encodes the local features to obtain their latent vector representations. Then, the mentions are ranked according to their disambiguation difficulty, which is measured by the learned vector representations. In order to enforce global coherence between mentions, Global Encoder encodes the local representations of mention-entity pairs in a sequential manner via a LSTM network, which maintains a long-term memory on features of entities which has been selected in previous states. Entity Selector uses a policy network to choose the target entities from the candidate set. For a single disambiguation decision, the policy network not only considers the pairs of current mention-entity representations, but also concerns the features of referred entities in the previous states which is pursued by the Global Encoder. In this way, Entity Selector is able to take actions based on the current state and previous ones. When eliminating the ambiguity of all mentions in the sequence, delayed rewards are used to adjust its policy in order to gain an optimized global decision.

Deep RL model, which learns to directly optimize the overall evaluation metrics, works much better than models which learn with loss functions that just evaluate a particular single decision. By this property, RL has been successfully used in many NLP tasks, such as information retrieval \cite{NogueiraC17}, dialogue system \cite{DhingraLLGCAD17} and relation classification \cite{FengHZYZ18}, etc. To the best of our knowledge, we are the first to design a RL model for global entity linking. And in this paper, our RL model is able to produce more accurate results by exploring the long-term influence of independent decisions and encoding the entities disambiguated in previous states.

In summary, the main contributions of our paper mainly include following aspects:
\begin{itemize}

\item We are the first to consider EL as a sequence decision problem and innovatively utilize a deep reinforcement learning model in this task.

\item The proposed model takes into account both local context and global coherence. In the process of global disambiguation, we make full use of the previous selected entity information and make decisions from a global perspective.

\item We evaluate our model on several benchmark datasets and the experimental results showed that our model achieves significant improvements over the state-of-the-art methods.
\end{itemize}

\begin{figure*}[t]
\centering
\includegraphics[width=6.8in]{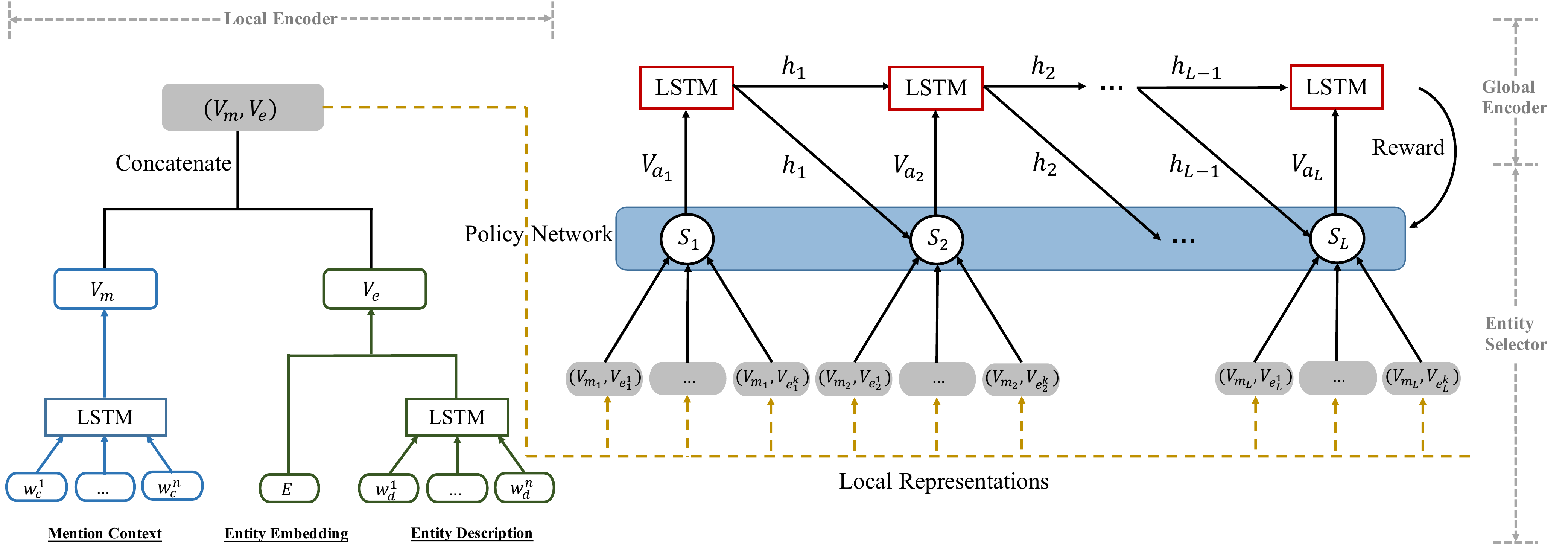}
\caption{The overall structure of our RLEL model. It contains three parts: Local Encoder, Global Encoder and Entity Selector. In this framework, $(V_{m_t}, V_{e_t^k})$ denotes the concatenation of the mention context vector $V_{m_t}$ and one candidate entity vector $V_{e_t^k}$. The policy network selects one entity from the candidate set, and $V_{a_t}$ denotes the concatenation of the mention context vector $V_{m_t}$ and the selected entity vector $V_{e_t^*}$. $h_t$ represents the hidden status of $V_{a_t}$, and it will be input into $S_{t+1}$.}

\end{figure*}

\section{Methodology}
The overall structure of our RLEL model is shown in Figure 2. The proposed framework mainly includes three parts: Local Encoder which encodes local features of mentions and their candidate entities, Global Encoder which encodes the global coherence of mentions in a sequence manner and Entity Selector which selects an entity from the candidate set. As the Entity Selector and the Global Encoder are correlated mutually, we train them jointly. Moreover, the Local Encoder as the basis of the entire framework will be independently trained before the joint training process starts. In the following, we will introduce the technical details of these modules.

\subsection{Preliminaries}
Before introducing our model, we firstly define the entity linking task. Formally, given a document $D$ with a set of mentions $M = \{m_1, m_2,...,m_k\}$, each mention $ m_t \in D$ has a set of candidate entities $C_{m_t} = \{e_{t}^1, e_{t}^2,..., e_{t}^n\}$. The task of entity linking is to map each mention $m_t$ to its corresponding correct target entity $e_{t}^+$ or return "NIL" if there is not correct target entity in the knowledge base. Before selecting the target entity, we need to generate a certain number of candidate entities for model selection.

Inspired by the previous works \cite{PappuBMST17, ZwicklbauerSG16, PhanSTHL17}, we use the mention's redirect and disambiguation pages in Wikipedia to generate candidate sets. For those mentions without corresponding disambiguation pages, we use its n-grams to retrieve the candidates \cite{PhanSTHL17}. In most cases, the disambiguation page contains many entities, sometimes even hundreds. To optimize the model's memory and avoid unnecessary calculations, the candidate sets need to be filtered \cite{0002HLL18, GaneaH17, TitovL18a}. Here we utilize the XGBoost model \cite{ChenG16} as an entity ranker to reduce the size of candidate set. The features used in XGBoost can be divided into two aspects, the one is string similarity like the Jaro-Winkler distance between the entity title and the mention, the other is semantic similarity like the cosine distance between the mention context representation and the entity embedding. Furthermore, we also use the statistical features based on the pageview and hyperlinks in Wikipedia. Empirically, we get the pageview of the entity from the Wikipedia Tool Labs\footnote{The url of the website is: https://tools.wmflabs.org/pageviews/} which counts the number of visits on each entity page in Wikipedia. After ranking the candidate sets based on the above features, we take the top k scored entities as final candidate set for each mention.

\subsection{Local Encoder}
Given a mention $m_t$ and the corresponding candidate set $\{e_t^1, e_t^2,..., \\ e_t^k\}$, we aim to get their local representation based on the mention context and the candidate entity description. For each mention, we firstly select its $n$ surrounding words, and represent them as word embedding using a pre-trained lookup table \cite{abs-1301-3781}. Then, we use Long Short-Term Memory (LSTM) networks to encode the contextual word sequence $\{w_c^1, w_c^2,..., w_c^n\}$ as a fixed-size vector $V_{m_t}$. The description of entity is encoded as $D_{e_t^i}$ in the same way. Apart from the description of entity, there are many other valuable information in the knowledge base. To make full use of these information, many researchers trained entity embeddings by combining the description, category, and relationship of entities. As shown in \cite{GaneaH17}, entity embeddings compress the semantic meaning of entities and drastically reduce the need for manually designed features or co-occurrence statistics. Therefore, we use the pre-trained entity embedding $E_{e_t^i}$ and concatenate it with the description vector $D_{e_t^i}$ to enrich the entity representation. The concatenation result is denoted by $V_{e_t^i}$.

After getting $V_{e_t^i}$, we concatenate it with $V_{m_t}$ and then pass the concatenation result to a multilayer perceptron (MLP). The MLP outputs a scalar to represent the local similarity between the mention $m_t$ and the candidate entity $e_t^i$. The local similarity is calculated by the following equations:

\begin{equation} 
\Psi(m_t, e_t^i) = MLP(V_{m_t}\oplus{V_{e_t^i}})
\end{equation}
Where $\oplus$ indicates vector concatenation. With the purpose of distinguishing the correct target entity and wrong candidate entities when training the local encoder model, we utilize a hinge loss that ranks ground truth higher than others. The rank loss function is defined as follows:

\begin{equation}       
L_{local} = max(0, \gamma-\Psi(m_t, e_t^+)+\Psi(m_t, e_t^-))
\end{equation}
When optimizing the objective function, we minimize the rank loss similar to \cite{GaneaH17, TitovL18a}. In this ranking model, a training instance is constructed by pairing a positive target entity $e_t^+$ with a negative entity $e_t^-$. Where $\gamma > 0$ is a margin parameter and our purpose is to make the score of the positive target entity $e_t^+$ is at least a margin $\gamma$ higher than that of negative candidate entity $e_t^-$.

With the local encoder, we obtain the representation of mention context and candidate entities, which will be used as the input into the global encoder and entity selector. In addition, the similarity scores calculated by MLP will be utilized for ranking mentions in the global encoder.

\subsection{Global Encoder}
In the global encoder module, we aim to enforce the topical coherence among the mentions and their target entities. So, we use an LSTM network which is capable of maintaining the long-term memory to encode the ranked mention sequence. What we need to emphasize is that our global encoder just encode the mentions that have been disambiguated by the entity selector which is denoted as $V_{a_t}$.

As mentioned above, the mentions should be sorted according to their contextual information and topical coherence.  So, we firstly divide the adjacent mentions into a segment by the order they appear in the document based on the observation that the topical consistency attenuates along with the distance between the mentions. Then, we sort mentions in a segment based on the local similarity and place the mention that has a higher similarity value in the front of the sequence. In Equation 1, we define the local similarity of $m_i$ and its corresponding candidate entity $e_t^i$. On this basis, we define $\Psi_{max}(m_i, e_i^a)$ as the the maximum local similarity between the $m_i$ and its candidate set $C_{m_i} = \{e_i^1, e_i^2,..., e_i^n\}$. We use $\Psi_{max}(m_i, e_i^a)$ as criterion when sorting mentions. For instance, if $\Psi_{max}(m_i, e_i^a) > \Psi_{max}(m_j, e_j^b)$ then we place $m_i$ before $m_j$. Under this circumstances, the mentions in the front positions may not be able to make better use of global consistency, but their target entities have a high degree of similarity to the context words, which allows them to be disambiguated without relying on additional information. In the end, previous selected target entity information is encoded by global encoder and the encoding result will be served as input to the entity selector.

Before using entity selector to choose target entities, we pre-trained the global LSTM network. During the training process, we input not only positive samples but also negative ones to the LSTM. By doing this, we can enhance the robustness of the network. In the global encoder module, we adopt the following cross entropy loss function to train the model.
\begin{equation}       
L_{global} = -\frac{1}{n}\sum_x{\left[y\ln{y^{'}} + (1-y)\ln(1-y^{'})\right]}
\end{equation}
Where $y\in\{0,1\}$ represents the label of the candidate entity. If the candidate entity is correct $y=1$, otherwise $y=0$. $y^{'}\in(0,1)$ indicates the output of our model. After pre-training the global encoder, we start using the entity selector to choose the target entity for each mention and encode these selections.

\begin{figure}[t]
\centering
\includegraphics[width=3.2in]{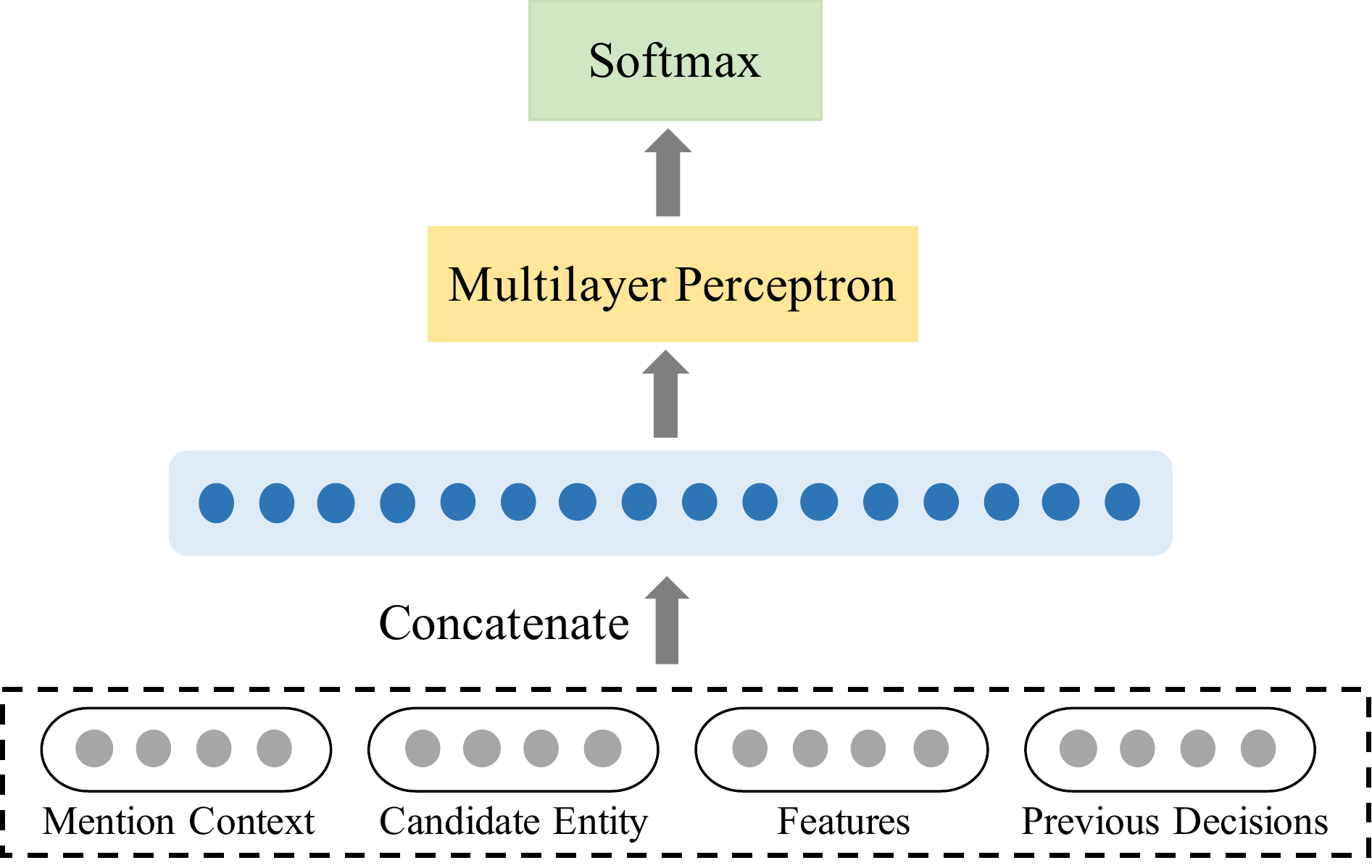}
\caption{The architecture of policy network. It is a feedforward neural network and the input consists of four parts: mention context representation, candidate entity representation, feature representation, and encoding of the previous decisions.}
\end{figure}

\subsection{Entity Selector}
In the entity selector module, we choose the target entity from candidate set based on the results of local and global encoder. In the process of sequence disambiguation, each selection result will have an impact on subsequent decisions. Therefore, we transform the choice of the target entity into a reinforcement learning problem and view the entity selector as an agent. In particular, the agent is designed as a policy network which can learn a stochastic policy and prevents the agent from getting stuck at an intermediate state \cite{XiongHW17}. Under the guidance of policy, the agent can decide which action (choosing the target entity from the candidate set)should be taken at each state, and receive a delay reward when all the selections are made. In the following part, we first describe the state, action and reward. Then, we detail how to select target entity via a policy network.

\subsubsection*{State}
The result of entity selection is based on the current state information. For time $t$, the state vector $S_t$ is generated as follows:
\begin{equation}       
S_t = V_{m_i}^t\oplus{V_{e_i}^t}\oplus{V_{feature}^t}\oplus{V_{e^*}^{t-1}}
\end{equation}
Where $\oplus$ indicates vector concatenation. The $V_{m_i}^t$ and $V_{e_i}^t$ respectively denote the vector of $m_i$ and $e_i$ at time $t$. For each mention, there are multiple candidate entities correspond to it. With the purpose of comparing the semantic relevance between the mention and each candidate entity at the same time, we copy multiple copies of the mention vector. Formally, we extend $V_{m_i}^t \in \mathbb{R}^{1\times{n}}$ to $V_{m_i}^t{'} \in \mathbb{R}^{k\times{n}}$ and then combine it with $V_{e_i}^t \in \mathbb{R}^{k\times{n}}$. Since $V_{m_i}^t$ and $V_{e_i}^t$ are mainly to represent semantic information, we add feature vector $V_{feature}^t$ to enrich lexical and statistical features. These features mainly include the popularity of the entity, the edit distance between the entity description and the mention context, the number of identical words in the entity description and the mention context etc. After getting these feature values, we combine them into a vector and add it to the current state. In addition, the global vector $V_{e^*}^{t-1}$ is also added to $S_t$. As mentioned in global encoder module, $V_{e^*}^{t-1}$ is the output of global LSTM network at time $t-1$, which encodes the mention context and target entity information from $m_0$ to $m_{t-1}$. Thus, the state $S_t$ contains current information and previous decisions, while also covering the semantic representations and a variety of statistical features. Next, the concatenated vector will be fed into the policy network to generate action.

\subsubsection*{Action}
According to the status at each time step, we take corresponding action. Specifically, we define the action at time step $t$ is to select the target entity $e_t^*$ for $m_t$. The size of action space is the number of candidate entities for each mention, where $a_i \in \{0,1,2...k\}$ indicates the position of the selected entity in the candidate entity list. Clearly, each action is a direct indicator of target entity selection in our model. After completing all the actions in the sequence we will get a delayed reward.


\subsubsection*{Reward}
The agent takes the reward value as the feedback of its action and learns the policy based on it. Since current selection result has a long-term impact on subsequent decisions, we don't give an immediate reward when taking an action. Instead, a delay reward is given by follows, which can reflect whether the action improves the overall performance or not.
\begin{equation}       
R(a_t) = p(a_t)\sum_{j=t}^{T}p(a_j) + (1 - p(a_t))(\sum_{j=t}^{T}p(a_j) + t - T)
\end{equation}
where $p(a_t)\in\{0,1\}$ indicates whether the current action is correct or not. When the action is correct $p(a_t)=1$ otherwise $p(a_t)=0$. Hence $\sum_{j=t}^{T}p(a_j)$ and $\sum_{j=t}^{T}p(a_j) + t - T$ respectively represent the number of correct and wrong actions from time t to the end of episode. Based on the above definition, our delayed reward can be used to guide the learning of the policy for entity linking.

\subsubsection*{Policy Network}
After defining the state, action, and reward, our main challenge becomes to choose an action from the action space. To solve this problem, we sample the value of each action by a policy network $\pi_{\Theta}(a|s)$. The structure of the policy network is shown in Figure 3. The input of the network is the current state, including the mention context representation, candidate entity representation, feature representation, and encoding of the previous decisions. We concatenate these representations and fed them into a multilayer perceptron, for each hidden layer, we generate the output by:
\begin{equation}       
h_i(S_t) = Relu(W_i*h_{i-1}(S_t) + b_i)
\end{equation}
Where $W_i$ and $ b_i$ are the parameters of the $i$th hidden layer, through the $relu$ activation function we get the $h_i(S_t)$. After getting the output of the last hidden layer, we feed it into a softmax layer which generates the probability distribution of actions. The probability distribution is generated as follows:
\begin{equation}       
\pi(a|s) = Softmax(W * h_l(S) + b)
\end{equation}
Where the $W$ and $b$ are the parameters of the softmax layer. For each mention in the sequence, we will take action to select the target entity from its candidate set. After completing all decisions in the episode, each action will get an expected reward and our goal is to maximize the expected total rewards. Formally, the objective function is defined as:
\begin{equation} 
\begin{split}      
J(\Theta) &= \mathbb{E}_{(s_t, a_t){\sim}P_\Theta{(s_t, a_t)}}R(s_1{a_1}...s_L{a_L}) \\
&=\sum_{t}\sum_{a}\pi_{\Theta}(a|s)R(a_t)
\end{split}
\end{equation}
Where $P_\Theta{(s_t, a_t)}$ is the state transfer function, $\pi_{\Theta}(a|s)$ indicates the probability of taking action $a$ under the state $s$, $R(a_t)$ is the expected reward of action $a$ at time step $t$. According to REINFORCE policy gradient algorithm\cite{Williams92}, we update the policy gradient by the way of equation 9. 
\begin{equation}
\Theta \leftarrow \Theta + \alpha \sum_{t}R(a_t)\nabla_{\Theta}\log\pi_{\Theta}(a|s)
\end{equation}
As the global encoder and the entity selector are correlated mutually, we train them jointly after pre-training the two networks. The details of the joint learning are presented in Algorithm 1.
 
\begin{algorithm}[t]
\caption{The Policy Learning for Entity Selector}
\begin{algorithmic}[1]
\REQUIRE Training data include multiple documents $D = \{D_1, D_2, ..., D_N\}$
\ENSURE The target entity for mentions $\Gamma = \{T_1, T_2, ..., T_N\}$ 

\STATE Initialize the policy network parameter $\Theta$, global LSTM network parameter $\Phi$;
\FOR{$D_k$ in $D$}
\STATE Generate the candidate set for each mention
\STATE Divide the mentions in $D_k$ into multiple sequences $S = \{S_1, S_2, ..., S_N\}$;
\FOR{$S_k$ in $S$}
\STATE Rank the mentions $M = \{m_1, m_2, ..., m_n\}$ in $S_k$ based on the local similarity;
\FOR{$m_k$ in $M$}
\STATE Sample the target entity $e_{k}^*$ for $m_k$ with $\Theta$;
\STATE Input the $V_{m_k}^t$ and ${V_{e_k^*}^t}$ to global LSTM network;
\ENDFOR
\STATE $//$ End of sampling, update parameters 
\STATE Compute delayed reward $R(a_t)$ for each action;
\STATE Update the parameter ${\Theta}^{'}$ of policy network:\\
\qquad $\Theta \leftarrow \Theta + \alpha \sum_{t}R(a_t)\nabla_{\Theta}\log\pi_{\Theta}(a|s)$

\ENDFOR
\STATE Update the parameter $\Phi$ in the global LSTM network\\
\ENDFOR
\end{algorithmic}
\end{algorithm}

\section{Experiment}
In order to evaluate the effectiveness of our method, we train the RLEL model and validate it on a series of popular datasets that are also used by \cite{GaneaH17, TitovL18a}. To avoid overfitting with one dataset, we use both  AIDA-Train and Wikipedia data in the training set. Furthermore, we compare the RLEL with some baseline methods, where our model achieves the state-of-the-art results. We implement our models in Tensorflow and run experiments on 4 Tesla V100 GPU. 


\subsection{Experiment Setup}
\subsubsection*{Datasets}
We conduct experiments on several different types of public datasets including news and encyclopedia corpus. The training set is AIDA-Train and Wikipedia datasets, where AIDA-Train contains 18448 mentions and Wikipedia contains 25995 mentions. In order to compare with the previous methods, we evaluate our model on AIDA-B and other datasets. These datasets are well-known and have been used for the evaluation of most entity linking systems. The statistics of the datasets are shown in Table 1.

\begin{itemize}
\item AIDA-CoNLL \cite{HoffartYBFPSTTW11} is annotated on Reuters news articles. It contains training (AIDA-Train), validation (AIDA-A) and test (AIDA-B) sets.
\item ACE2004 \cite{RatinovRDA11} is a subset of the ACE2004 Coreference documents.
\item MSNBC \cite{Cucerzan07} contains top  two  stories  in  the  ten news categories(Politics, Business, Sports etc.)
\item AQUAINT \cite{MilneW08} is a news corpus from the Xinhua News Service, the New York Times, and the Associated Press.
\item WNED-CWEB \cite{GuoB18} is randomly picked from the FACC1 annotated ClueWeb 2012 dataset.
\item WNED-WIKI \cite{GuoB18} is crawled from Wikipedia pages with its original hyperlink annotation.
\item OURSELF-WIKI is crawled by ourselves from Wikipedia pages.
\end{itemize} 

\begin{table}
\caption{Statistics of document and mention numbers on experimental datasets.}
\renewcommand\arraystretch{1.1}
\newcommand{\tabincell}[2]{\begin{tabular}{@{}#1@{}}#2\end{tabular}}
\centering
\scalebox{0.9}{
\begin{tabular}{|c|c|c|c|} 
\hline 
{Dataset}  & {Doc Num} & {Mention Num} & {Mentions Per Doc}\\
\hline
AIDA-Train  & 946 & 18448 & 19.5  \\  
AIDA-A  & 216 & 4791 & 22.1  \\
AIDA-B  & 231 & 4485 & 19.4  \\
\hline 
ACE2004  & 36 & 251 & 7.1  \\
\hline 
MSNBC  & 20 & 656 & 32.8  \\
\hline 
AQUAINT  & 50 & 727 & 14.5  \\
\hline 
WNED-CWEB  & 320 & 11154 & 34.8  \\
\hline 
WNED-WIKI  & 320 & 6821 & 21.3  \\
\hline 
OURSELF-WIKI  & 460 & 25995 & 56.5  \\
\hline 
\end{tabular}
}
\setlength{\abovecaptionskip}{8pt}
\end{table}

\subsubsection*{Training Details}
During the training of our RLEL model, we select top K candidate entities for each mention to optimize the memory and run time. In the top K candidate list, we define the recall of correct target entity is $R_t$. According to our statistics, when K is set to 1, $R_t$ is 0.853, when K is 5, $R_t$ is 0.977, when K increases to 10, $R_t$ is 0.993. Empirically, we choose top 5 candidate entities as the input of our RLEL model. For the entity description, there are lots of redundant information in the wikipedia page, to reduce the impact of noise data, we use TextRank algorithm \cite{MihalceaT04} to select 15 keywords as description of the entity. Simultaneously, we choose 15 words around mention as its context. In the global LSTM network, when the number of mentions does not reach the set length, we adopt the mention padding strategy. In short, we copy the last mention in the sequence until the number of mentions reaches the set length.

\subsubsection*{Hyper-parameter setting}
We set the dimensions of word embedding and entity embedding to 300, where the word embedding and entity embedding are released by \cite{PenningtonSM14} and \cite{GaneaH17} respectively. For parameters of the local LSTM network, the number of LSTM cell units is set to 512, the batch size is 64, and the rank margin $\gamma$ is 0.1. Similarly, in global LSTM network, the number of LSTM cell units is 700 and the batch size is 16. In the above two LSTM networks, the learning rate is set to 1e-3,  the probability of dropout is set to 0.8, and the Adam is utilized as optimizer. In addition, we set the number of MLP layers to 4 and extend the priori feature dimension to 50 in the policy network.

\subsection{Comparing with Previous Work}
\subsubsection*{Baselines}
We compare RLEL with a series of EL systems which report state-of-the-art results on the test datasets. There are various methods including classification model \cite{MilneW08}, rank model \cite{ChisholmH15, RatinovRDA11} and probability graph model \cite{GuoB18, HoffartYBFPSTTW11, HuangHJ15, GaneaH17, TitovL18a}. Except that, Cheng $et$ $al.$\cite{ChengR13} formulate their global decision problem as an Integer Linear Program (ILP) which incorporates the entity-relation inference. Globerson $et$ $al.$ \cite{GlobersonLCSRP16} introduce a multi-focal attention model which allows each candidate to focus on limited mentions, Yamada $et$ $al.$\cite{YamadaS0T16} propose a word and entity embedding model specifically designed for EL. 

\subsubsection*{Evaluation Metric}
We use the standard Accuracy, Precision, Recall and F1 at mention level (Micro) as the evaluation metrics:
\begin{equation}
Accuracy = \frac{|M \cap M^*|}{|M \cup M^*|}
\end{equation}

\begin{equation}
Precision = \frac{|M \cap M^*|}{|M|}
\end{equation}

\begin{equation}
Recall = \frac{|M \cap M^*|}{|M^*|}
\end{equation}

\begin{equation}
F1 = \frac{2*Precision*Recall}{Precision+Recall}
\end{equation}
where $M^*$ is the golden standard set of the linked name mentions, $M$ is the set of linked name mentions outputted by an EL method.

\begin{table}[tp]
\caption{In-KB accuracy result on AIDA-B dataset.}
\renewcommand\arraystretch{1}
\newcommand{\tabincell}[2]{\begin{tabular}{@{}#1@{}}#2\end{tabular}}
\centering
\begin{tabular}{|c|c|} 
\hline  
{Methods}  & {AIDA-B} \\
\hline  
Huang and Heck (2015)\cite{HuangHJ15}  & 86.6\%  \\
\hline  
Chisholm and Hachey (2015)\cite{ChisholmH15}  & 88.7\%  \\
\hline 
Guo and Barbosa (2016)\cite{GuoB18}  & 89.0\%  \\
\hline 
Globerson \emph{et al.} (2016)\cite{GlobersonLCSRP16}  & 91.0\%  \\
\hline 
Yamada \emph{et al.} (2016)\cite{YamadaS0T16}  & 91.5\%  \\
\hline 
Ganea and Hofmann (2017)\cite{GaneaH17}  & 92.2\%  \\
\hline 
Phong and Titov (2018)\cite{TitovL18a}  & 93.1\%  \\
\hline
our  & {\bfseries 94.3\%}  \\
\hline  
\end{tabular}
\setlength{\abovecaptionskip}{8pt}
\end{table}

\begin{table*}[tp]
\caption{Compare our model with other baseline methods on different types of datasets. The evaluation metric is micro F1.
}
\renewcommand\arraystretch{1}
\newcommand{\tabincell}[2]{\begin{tabular}{@{}#1@{}}#2\end{tabular}}
\centering
\begin{tabular}{|c|c|c|c|c|c|c|} 
\hline
{Methods}  & {MSNBC}  & {AQUAINT}  & {ACE2004}  & {CWEB}  & {WIKI}& {Avg} \\
\hline  
Milne and Witten (2008)\cite{MilneW08}  & 78\%  & 85\%  &  81\%  & 64.1\%  &  81.7\%  & 77.96\% \\
\hline 
Hoffart and Johannes(2011)\cite{HoffartYBFPSTTW11}  & 79\%  & 56\%  & 80\%  & 58.6\%  & 63\%  & 67.32\% \\
\hline 
Ratinov and Lev\cite{RatinovRDA11}   & 75\%  & 83\%  & 82\%  & 56.2\%  & 67.2\%  & 72.68\% \\
\hline 
Cheng and Roth (2013)\cite{ChengR13}  & 90\%  & {\bfseries 90\%}  & 86\%  & 67.5\%  & 73.4\%  & 81.38\% \\
\hline 
Guo and Barbosa (2016)\cite{GuoB18}  & 92\%  & 87\%  & 88\%  & 77\%  & {\bfseries 84.5\%}  & 85.7\% \\
\hline 
Ganea and Hofmann (2017)\cite{GaneaH17}  & 93.7\%  & 88.5\%  & 88.5\%  & 77.9\%  & 77.5\%  & 85.22\% \\
\hline 
Phong and Titov (2018)\cite{TitovL18a}  & {\bfseries 93.9\%}  & 88.3\%  & 89.9\%  & 77.5\% & 78.0\%  & 85.51\% \\
\hline
our  & 92.8\%   & 87.5\%  & {\bfseries 91.2\%}  & {\bfseries 78.5\%}  &  82.8\%  & {\bfseries 86.56\%}  \\
\hline

\end{tabular}
\setlength{\abovecaptionskip}{8pt}
\end{table*} 

\subsubsection*{Results}
Same as previous work, we use in-KB accuracy and micro F1 to evaluate our method. We first test the model on the AIDA-B dataset. From Table 2, we can observe that our model achieves the best result. Previous best results on this dataset are generated by \cite{GaneaH17, TitovL18a} which both built CRF models. They calculate the pairwise scores between all candidate entities. Differently, our model only considers the consistency of the target entities and ignores the relationship between incorrect candidates. The experimental results show that our model can reduce the impact of noise data and improve the accuracy of disambiguation. Apart from experimenting on AIDA-B, we also conduct experiments on several different datasets to verify the generalization performance of our model. 

From Table 3, we can see that RLEL has achieved relatively good performances on ACE2004, CWEB and WIKI. At the same time, previous models \cite{GaneaH17, TitovL18a, ChengR13} achieve better performances on the news datasets such as MSNBC and AQUINT, but their results on encyclopedia datasets such as WIKI are relatively poor. To avoid overfitting with some datasets and improve the robustness of our model, we not only use AIDA-Train but also add Wikipedia data to the training set. In the end, our model achieve the best overall performance.

For most existing EL systems, entities with lower frequency are difficult to disambiguate. To gain further insight, we analyze the accuracy of the AIDA-B dataset for situations where gold entities have low popularity. We divide the gold entities according to their pageviews in wikipedia, the statistical disambiguation results are shown in Table 4. Since some pageviews can not be obtained, we only count part of gold entities. The result indicates that our model is still able to work well for low-frequency entities. But for medium-frequency gold entities, our model doesn't work well enough. The most important reason is that other candidate entities corresponding to these medium-frequency gold entities have higher pageviews and local similarities, which makes the model difficult to distinguish.

\begin{table}
\caption{The micro F1 of gold entities with different pageviews on part of AIDA-B dataset.}
\renewcommand\arraystretch{1}
\newcommand{\tabincell}[2]{\begin{tabular}{@{}#1@{}}#2\end{tabular}}
\centering
\scalebox{1.1} {
\begin{tabular}{|c|c|c|}
\hline  
{Pageview/million} & {Mention Num} & {Micro F1} \\
\hline  
$<$ 0.01 &307 & 91.93\%  \\
\hline  
0.01-0.1  &612 &  86.06\%  \\
\hline 
0.1-1  &968 & 88.97\%  \\
\hline 
1-5 &1006 & 96.03\%  \\
\hline 
5-10  &493 & 96.43\%  \\
\hline 
$>$ 10  &825 & 99.39\%  \\
\hline 
\end{tabular}
}
\setlength{\abovecaptionskip}{8pt}
\end{table}

\subsection{Discussion on different RLEL variants}
To demonstrate the effects of RLEL, we evaluate our model under different conditions. First, we evaluate the effect of sequence length on global decision making. Second, we assess whether sorting the mentions have a positive effect on the results. Third, we analysis the results of not adding globally encoding during entity selection. Last, we compare our RL selection strategy with the greedy choice. 

\subsubsection*{Sequence in different length}
A document may contain multiple topics, so we do not add all mentions to a single sequence. In practice, we add some adjacent mentions to the sequence and use reinforcement learning to select entities from beginning to end. To analysis the impact of the number of mentions on joint disambiguation, we experiment with sequences on different lengths. The results on AIDA-B are shown in Figure 4. We can see that when the sequence is too short or too long, the disambiguation results are both very poor. When the sequence length is less than 3, delay reward can't work in reinforcement learning, and when the sequence length reaches 5 or more, noise data may be added. Finally, we choose the 4 adjacent mentions to form a sequence.

\begin{figure}[t]
\centering
\includegraphics[width=2.4in]{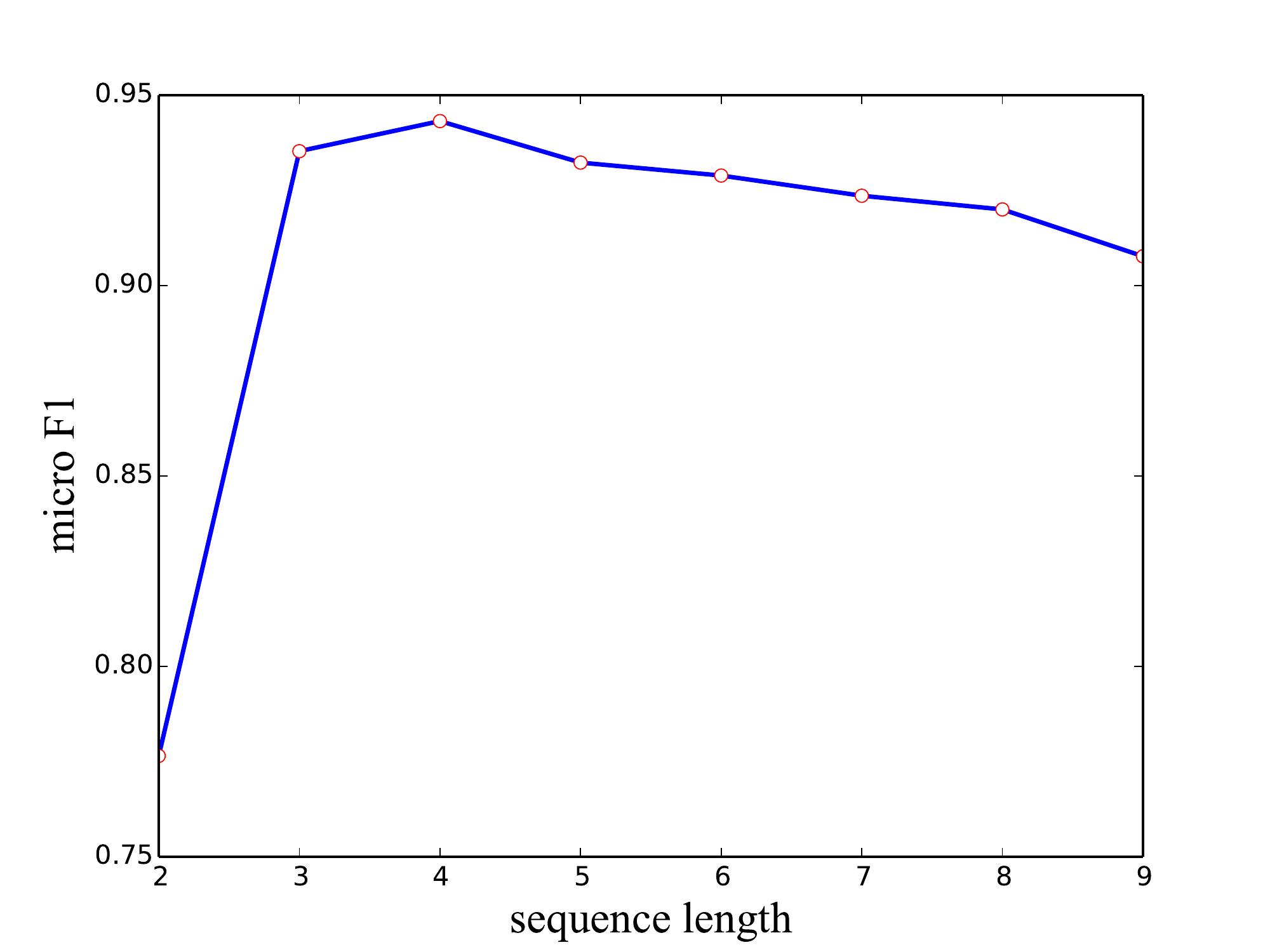}
\caption{The performance of models with different sequence lengths on AIDA-B dataset.}
\end{figure}

\begin{figure*}[t]
\centering
\includegraphics[width=7in]{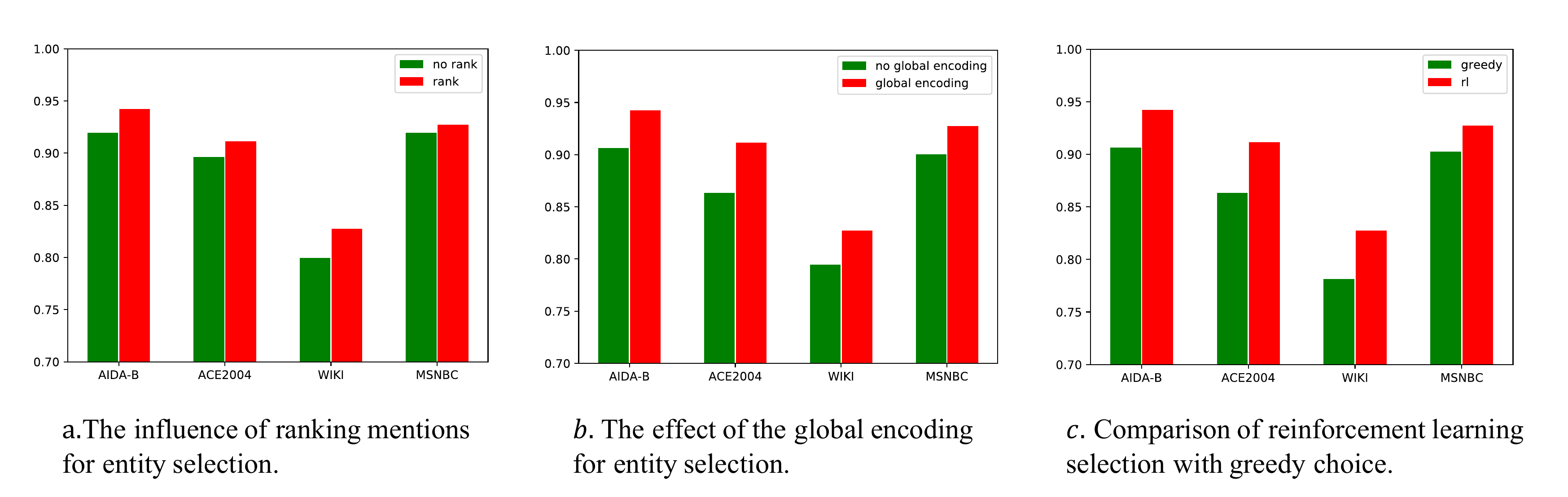}
\caption{The comparative experiments of RLEL model.}
\end{figure*}

\begin{table*}[t]
\caption{Entity selection examples by our RLEL model.}
\renewcommand\arraystretch{1}
\newcommand{\tabincell}[2]{\begin{tabular}{@{}#1@{}}#2\end{tabular}}
\centering
\begin{tabular}{|l|l|l|} 
\hline
{Document Content}  & {Mentions after ranking}  & {Selected Target Entity(is correct)} \\
\hline
\tabincell{l}{{\bfseries Australia} beat {\bfseries West Indies} by five wickets \\ in a {\bfseries World Series} limited overs match at the \\ {\bfseries Melbourne Cricket Ground}on Friday...} & \tabincell{l}{1.Melbourne Cricket Ground \\ 2.World Series \\ 3.West Indies \\ 4.Australia} & \tabincell{l}{1.Melbourne Cricket Ground(correct) \\ 2.World Series Cricket(correct) \\ 3.West Indies cricket team(correct)\\ 4.Australia national cricket team(correct)} \\

\hline 
\tabincell{l}{Instead of {\bfseries Los Angeles International}, \\..., consider flying into {\bfseries Burbank} or {\bfseries John} \\ {\bfseries Wayne Airport} in {\bfseries Orange County}, Calif...}  & \tabincell{l}{1.John Wayne Airport \\ 2.Orange County \\ 3.Los Angeles International \\ 4.Burbank}  & \tabincell{l}{1.John Wayne Airport(correct) \\ 2.Orange County, California(correct) \\ 3.Los Angeles International Airport(correct) \\ 4.Burbank, California\\(wrong, the correct is "Hollywood Burbank Airport")}\\
\hline
\end{tabular}
\setlength{\abovecaptionskip}{8pt}
\end{table*}   

\subsubsection*{Influence of ranking mentions}
In this section, we test whether ranking mentions is helpful for entity selections. At first, we directly input them into the global encoder by the order they appear in the text. We record the disambiguation results and compare them with the method which adopts ranking mentions. As shown in Figure 5a, the model with ranking mentions has achieved better performances on most of datasets, indicating that it is effective to place the mention that with a higher local similarity in front of the sequence. It is worth noting that the effect of ranking mentions is not obvious on the MSNBC dataset, the reason is that most of mentions in MSNBC have similar local similarities, the order of disambiguation has little effect on the final result.

\subsubsection*{Effect of global encoding}
Most of previous methods mainly use the similarities between entities to correlate each other, but our model associates them by encoding the selected entity information. To assess whether the global encoding contributes to disambiguation rather than add noise, we compare the performance with and without adding the global information. When the global encoding is not added, the current state only contains the mention context representation, candidate entity representation and feature representation, notably, the selected target entity information is not taken into account. From the results in Figure 5b, we can see that the model with global encoding achieves an improvement of 4\% accuracy over the method that without global encoding.

\subsubsection*{Different entity selection strategies}
To illustrate the necessity for adopting the reinforcement learning for entity selection, we compare two entity selection strategies like \cite{FengHZYZ18}. Specifically, we perform entity selection respectively with reinforcement learning and greedy choice. The greedy choice is to select the entity with largest local similarity from candidate set. But the reinforcement learning selection is guided by delay reward, which has a global perspective. In the comparative experiment, we keep the other conditions consistent, just replace the RL selection with a greedy choice. Based on the results in Figure 5c, we can draw a conclusion that our entity selector perform much better than greedy strategies.

\subsection{Case Study}
Table 5 shows two entity selection examples by our RLEL model. For multiple mentions appearing in the document, we first sort them according to their local similarities, and select the target entities in order by the reinforcement learning model. From the results of sorting and disambiguation, we can see that our model is able to utilize the topical consistency between mentions and make full use of the selected target entity information.  

\section{Related Work}
The related work can be roughly divided into two groups: entity linking and reinforcement learning.

\subsection{Entity Linking}
Entity linking falls broadly into two major approaches: local and global disambiguation. Early studies use local models to resolve mentions independently, they usually disambiguate mentions based on lexical matching between the mention's surrounding words and the entity profile in the reference KB. Various methods have been proposed to model mention's local context ranging from binary classification \cite{MilneW08} to rank models \cite{DredzeMRGF10, ChenJ11}. In these methods, a large number of hand-designed features are applied. For some marginal mentions that are difficult to extract features, researchers also exploit the data retrieved by search engines \cite{CornoltiFCSR14, CornoltiFCRS16} or Wikipedia sentences \cite{TanWRLZ17}. However, the feature engineering and search engine methods are both time-consuming and laborious. Recently, with the popularity of deep learning models, representation learning is utilized to automatically find semantic features \cite{GuptaSR17, CaoHJCL17}. The learned entity representations which by jointly modeling textual contexts and knowledge base are effective in combining multiple sources of information. To make full use of the information contained in representations, we also utilize the pre-trained entity embeddings in our model.

In recent years, with the assumption that the target entities of all mentions in a document shall be related, many novel global models for joint linking are proposed. Assuming the topical coherence among mentions, authors in \cite{GaneaGLEH16, RanSW18} construct factor graph models, which represent the mention and candidate entities as variable nodes, and exploit factor nodes to denote a series of features. Two recent studies \cite{GaneaH17, TitovL18a} use fully-connected pairwise Conditional Random Field(CRF) model and exploit loopy belief propagation to estimate the max-marginal probability. Moreover, PageRank or Random Walk \cite{HanSZ11, GuoB18, ZwicklbauerSG16} are utilized to select the target entity for each mention. The above probabilistic models usually need to predefine a lot of features and are difficult to calculate the max-marginal probability as the number of nodes increases. In order to automatically learn features from the data, Cao \emph{et al.} \cite{0002HLL18} applies Graph Convolutional Network to flexibly encode entity graphs. However, the graph-based methods are computationally expensive because there are lots of candidate entity nodes in the graph.

To reduce the calculation between candidate entity pairs, Globerson \emph{et al.} \cite{GlobersonLCSRP16} introduce a coherence model with an attention mechanism, where each mention only focus on a fixed number of mentions. Unfortunately, choosing the number of attention mentions is not easy in practice. Two recent studies \cite{PhanSTHL17, abs-1802-01074} finish linking all mentions by scanning the pairs of mentions at most once, they assume each mention only needs to be consistent with one another mention in the document. The limitation of their method is that the consistency information is too sparse, resulting in low confidence. Similar to us, Guo \emph{et al.} \cite{GuoB18} also sort mentions according to the difficulty of disambiguation, but they did not make full use of the information of previously referred entities for the subsequent entity disambiguation. Nguyen \emph{et al.} \cite{NguyenFRHGS16} use the sequence model, but they simply encode the results of the greedy choice, and measure the similarities between the global encoding and the candidate entity representations. Their model does not consider the long-term impact of current decisions on subsequent choices, nor does they add the selected target entity information to the current state to help disambiguation.

\subsection{Reinforcement Learning}
In the last few years, reinforcement learning has emerged as a powerful tool for solving complex sequential decision-making problems. It is well known for its great success in the game field, such as Go \cite{SilverHMGSDSAPL16} and Atari games \cite{MnihKSRVBGRFOPB15}. Recently, reinforcement learning has also been successfully applied to many natural language processing tasks and achieved good performance \cite{XiongHW17,LiuWZYZ18,FengHZYZ18}. Feng \emph{et al.}\cite{FengHZYZ18} used reinforcement learning for relation classification task by filtering out the noisy data from the sentence bag and they achieved huge improvements compared with traditional classifiers. Zhang \emph{et al.} \cite{ZhangHZ18} applied the reinforcement learning on sentence representation by automatically discovering task-relevant structures. To automatic taxonomy induction from a set of terms, Han \emph{et al.} \cite{HanRSMG18} designed an end-to-end reinforcement learning model to determine which term to select and where to place it on the taxonomy, which effectively reduced the error propagation between two phases. Inspired by the above works, we also add reinforcement learning to our framework.

\section{Conclusions}
In this paper we consider entity linking as a sequence decision problem and  present a reinforcement learning based model. Our model learns the policy on selecting target entities in a sequential manner and makes decisions based on current state and previous ones. By utilizing the information of previously referred entities, we can take advantage of global consistency to disambiguate mentions. For each selection result in the current state, it also has a long-term impact on subsequent decisions, which allows learned policy strategy has a global view. In experiments, we evaluate our method on AIDA-B and other well-known datasets, the results show that our system outperforms state-of-the-art solutions. In the future, we would like to use reinforcement learning to detect mentions and determine which mention should be firstly disambiguated in the document.

\begin{acks}

This research is supported by the \grantsponsor{GS501100001809}{National Key Research and Development Program of China }{}(No.~\grantnum{GS501100001809}{2018YFB1004703}),  \grantsponsor{GS501100001809}{the Beijing Municipal Science and Technology Project}{} under grant (No.~\grantnum{GS501100001809}{\\Z181100002718004}), and \grantsponsor{GS501100001809}{the National Natural Science Foundation of China}{} grants(No.~\grantnum{GS501100001809}{61602466}).
\end{acks}

%
\bibliographystyle{ACM-Reference-Format}
\bibliography{www_el.bib}

\end{document}